\documentclass[sigconf,authorversion]{acmart}
\usepackage{amsmath}
\renewcommand\footnotetextcopyrightpermission[1]{}
\makeatletter
\def\runningfoot{\def\@runningfoot{}}
\def\firstfoot{\def\@firstfoot{}}
\makeatother

\AtBeginDocument{%
  \providecommand\BibTeX{{%
    \normalfont B\kern-0.5em{\scshape i\kern-0.25em b}\kern-0.8em\TeX}}}

\usepackage{graphicx}
\usepackage{bbm}
\usepackage{dsfont}
\usepackage{amsmath}
\usepackage{threeparttable}
\usepackage[font=bf]{caption}
\usepackage{bm}
\usepackage{natbib}
\usepackage{etoolbox}
\usepackage{fancyhdr}

\DeclareCaptionLabelFormat{plain}{#1 #2}
\begin{document}

\fancypagestyle{main}{%
  \fancyhf{}
  \renewcommand{\headrulewidth}{0pt}
}

\pagestyle{fancy}
\fancyhf{}
\renewcommand{\headrulewidth}{0pt}

\title{A Teacher Is Worth A Million Instructions}
\author{
    \textbf{Nikhil Kothari} \quad \textbf{Ravindra Nayak} \quad \textbf{Shreyas Shetty}\\
    \textbf{Amey Patil} \quad \textbf{Nikesh Garera}\\[10pt]
    % {\fontsize{12.5}{30}\selectfont Flipkart AI}\\[5pt]
    \texttt{\{nikhil.kothari, ravindra.nayak, shreyas.shetty,}\\
    \texttt{amey.patil, nikesh.garera\}@flipkart.com}
}

\begin{abstract}
Large Language Models(LLMs) have shown exceptional abilities, yet training these models can be quite challenging. There is a strong dependence on the quality of data and finding the best instruction tuning set. Further, the inherent limitations in training methods create substantial difficulties to train relatively smaller models with 7B and 13B parameters. In our research, we suggest an improved training method for these models by utilising knowledge from larger models, such as a mixture of experts (8x7B) architectures. The scale of these larger models allows them to capture a wide range of variations from data alone, making them effective teachers for smaller models. Moreover, we implement a novel post-training domain alignment phase that employs domain-specific expert models to boost domain-specific knowledge during training while preserving the model's ability to generalise. Fine-tuning Mistral 7B and 2x7B with our method\footnote{Code can be found at \url{https://github.com/flipkart-incubator/kd-dae}} surpasses the performance of state-of-the-art language models with more than 7B and 13B parameters: achieving up to $7.9$ in MT-Bench \cite{zheng2023judging} and $93.04\%$ on AlpacaEval \cite{alpaca_eval}. 

% We release code\footnote{https://github.com/dawnik17/kd-dae} and model checkpoints for Flip-7B-Instruct-$\alpha$ (7B)\footnote{https://huggingface.co/dawn17/Flip-7B-Instruct-alpha} and Flip-2x7B-Instruct-$\alpha$ (13B)\footnote{https://huggingface.co/dawn17/Flip-2x7B-Instruct-alpha}.
\end{abstract}
\maketitle

\section{I\MakeLowercase{ntroduction}}
The recent advancements in Large Language Models (LLMs) have significantly propelled the field of natural language understanding and generation. Pre-trained language models (PLMs) leveraging extensive training corpus sourced from web \cite{together2023redpajama, pile} have demonstrated impressive capabilities across various natural language processing (NLP) tasks. However, additional training steps are required for PLMs to follow instructions and keep the responses aligned to human preferences. 

Instruction tuning (IT) \cite{alpaca, wei2022finetuned, zhang2024instruction} trains a PLM further for instruction following; utilising the general knowledge imparted in the pre-training phase along with the imparted instruction following capability it trains the model to generalise well on unseen tasks. However, while proficient at following instructions, these models may produce outputs that are potentially toxic or ethically questionable. To enhance alignment with human values, further training is necessary, utilizing techniques such as reinforcement learning with human feedback \cite{li2023reinforcement}, direct preference optimization (DPO) \cite{rafailov2023direct} and monolithic preference optimization without reference model (ORPO) \cite{hong2024orpo} based on pairwise preference data.

Instruction tuning requires meticulous attention to data quality, optimization of instruction sets, and the implementation of effective training methodologies to ensure peak performance. A primary challenge in training these instruction-tuned models in specific domains is the potential reduction in the model's generalization ability, a factor we monitor using public evaluation benchmarks in our research. In this study, we present a method that not only addresses these concerns but also improves public benchmarks while aligning the model within a specific domain, in this instance, e-commerce. Drawing from the successful implementation of Knowledge Distillation (KD) \cite{hinton2015distilling} in miniLLMs\cite{gu2023minillm} and tasks such as classification, we propose it as an alternative to the commonly used supervised fine-tuning (SFT) and alignment process in language model training. We propose {\it Domain Alignment from Expert (DAE)\/}, a unique post-training domain alignment algorithm designed to strengthen domain-specific knowledge within the LLMs. DAE integrates domain-specific expert models into the training process, enhancing the model's understanding of specialized domains while preserving its ability to generalize across broader contexts. This approach surpasses state-of-the-art language models with over 7B and 13B parameters, as evidenced by significant improvements in MT-Bench and AlpacaEval benchmarks.

\begin{figure}[h]
    \centering
    \includegraphics[width = 0.5\textwidth]{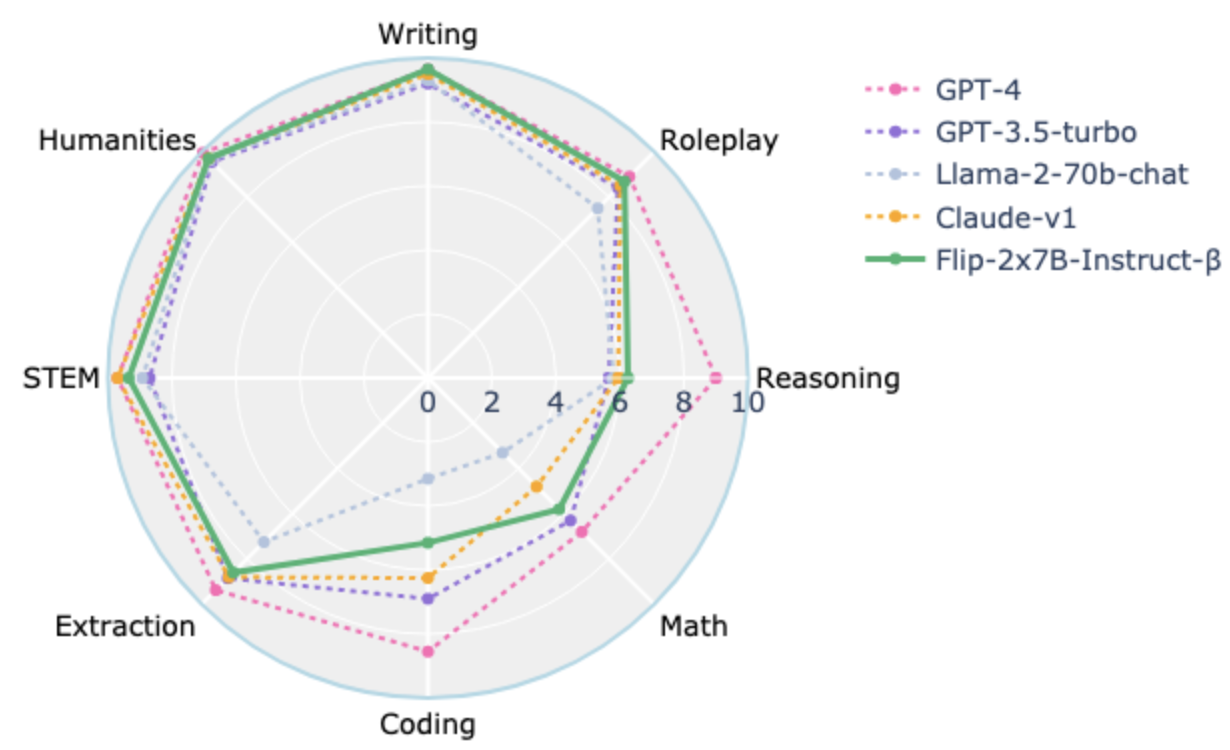}
    \caption{\textmd{Model performance on MT-Bench. We compare Flip-2x7B-Instruct, trained with KD and DAE, to proprietary as well as larger, open-access models like Llama-2-70B-chat.}}
    \label{fig:dae}
\end{figure}
% \vspace{-0.55cm}
Our main contribution is the successful imparting of knowledge from domain experts while retaining the generalization capabilities of the student model. This study challenges two commonly accepted beliefs: (i) KD with a teacher model smaller than the student model doesn't work, and (ii) KD from larger models cannot yield performance that is equal to or better than SFT and alignment combined. Our work aligns the model within the e-commerce domain, while demonstrating its generalization ability across tasks such as entity extraction, summarization, roleplay, and reasoning using language models with up to 13B parameters.

\section{M\MakeLowercase{ethod}}
In a decoder-only model employing self-attention \cite{vaswani2023attention}, the task involves generating an output sequence, where variables \(x_1, ..., x_n\) represent the input symbols, and \(z_1, ..., z_m\) denote the output symbols. The model learns to produce the symbol \(z_{t+1}\) at time-step \(t\), while only considering preceding symbols ( \(x_1, ..., x_n, z_1,..., z_{t}\) ).
During training, Cross-Entropy Loss (CE) is employed as the objective function at every step. CE for a given sequence quantifies the dissimilarity between the predicted probability distribution \( q_{\theta}(y|x) \), parameterized by \( \theta \), and the true probability distribution \( p(y|x) \) at each step in the sequence.

\begin{equation}
\text{CE}(p,q_{\theta}) = -\frac{1}{T-1} \sum_{t=1}^{T-1} p(y_{t+1}|x_{t' \leq t}) \log \left( q_{\theta}(y_{t+1}|x_{t' \leq t}) \right)
\end{equation}

Here, \( T \) denotes the length of the sequence, \( y_{t+1} \) is the label at \( {t+1} \) step, \( x_{t' \leq t} \) is the input sequence till step \( {t} \), \( p(y_{t+1}|x_{t' \leq t}) \) and \( q_{\theta}(y_i|x_{t' \leq t}) \) represent the respective true and predicted probability of \( y_{t+1} \) given \( x_{t' \leq t} \). In practice since the true probability of \( y_{t+1} \) is unknown, \( p(y_{t+1}|x_{t' \leq t}) \) is set to 1 for the \( y_{t+1} \) token and 0 for the rest of the tokens in vocabulary. This complete dependence on \( y_{t+1} \) makes the training data quality and its diversity paramount, and thus also presents a need for something more nuanced. Instead of considering the true probability as a one-hot vector, we could use a teacher for a better estimate of the distribution. 

\subsection{K\MakeLowercase{nowledge} D\MakeLowercase{istillation} in T\MakeLowercase{ransformer}}

The suggested method focuses on the transformers architecture. Our experiments are conducted on Mistral models, namely Mistral 7B v0.1 Base, Mistral Instruct v2, and Mixtral 8x7B Instruct. These models share the same tokenizer, number of decoder layers, attention heads, and hidden dimension, facilitating straightforward distillation with a one-to-one mapping. However, as long as the tokenizer of the teacher and student models is the same, we can still distill knowledge from the teacher with some adjustments. We only focus on the prediction layer and attention based distillation as a part of our study. Even though adding hidden states based distillation also could yield good results, we couldn't extensively try it out due to its high memory demands.

\textbf{Prediction Layer Distillation}. To fit the prediction of the teacher model, we use Kullback-Leibler Divergence (KLD) between the student model logits and the teacher model logits. KLD loss for a sequence measures the difference between the student probability distribution \( q_{\theta}(y|x) \) parameterized by \( \theta \) and the teacher probability distribution \( p(y|x) \) \cite{csiszar1975divergence} at every step.

\begin{equation}
\mathcal{L}_{\text{pred}} = \frac{1}{T} \sum_{t=1}^{T} \sum_{i=1}^{V} p(y_i|x_{t' \leq t}) \log \left( \frac{p(y_i|x_{t' \leq t})}{q_{\theta}(y_i|x_{t' \leq t})} \right)
\label{eq:pred}
\end{equation}

where \( T \) is the sequence length, \( V \) is the vocabulary size, \( y_i \) is the \( i^{\text{th}} \) token in the vocabulary, \( x_{t' \leq t} \) is the input sequence till step \( {t} \), \( p(y_i|x_{t' \leq t}) \) and \( q(y_i|x_{t' \leq t}) \) are the respective true (teacher) and the predicted (student) probabilities of \( y_i \)  given \( x_{t' \leq t} \). Here, \( p(y|x) \) is a distribution over the entire vocabulary and hence, gives us a smoother "true" distribution 
at every step while reducing the overt dependence on the training data quality.

\textbf{Attention Based Distillation}. In the transformer architecture, attention weights determine the importance of each token in relation to others at each time-step in the sequence. This attention mechanism enables the model to focus on relevant information, allowing for better capture of long-range dependencies. The Transformer model typically comprises multiple layers, each consisting of parallel self-attention heads. These heads allow the model to attend to different parts of the input simultaneously, facilitating efficient learning of contextual relationships. Mistral 7B v0.1 Base, Mistral 7B Instruct v2 and Mixtral 8x7B Instruct, all 3 have 32 decoder layers and 32 attention heads. This architectural similarity simplifies attention-based calculations.

For any time-step $t$, the attention weights satisfy the constraint:

\begin{equation}
\sum_{i=1}^{t} a_{it} = 1
\end{equation}

\begin{figure}[h]
    \centering
    \includegraphics[width=0.4\textwidth]{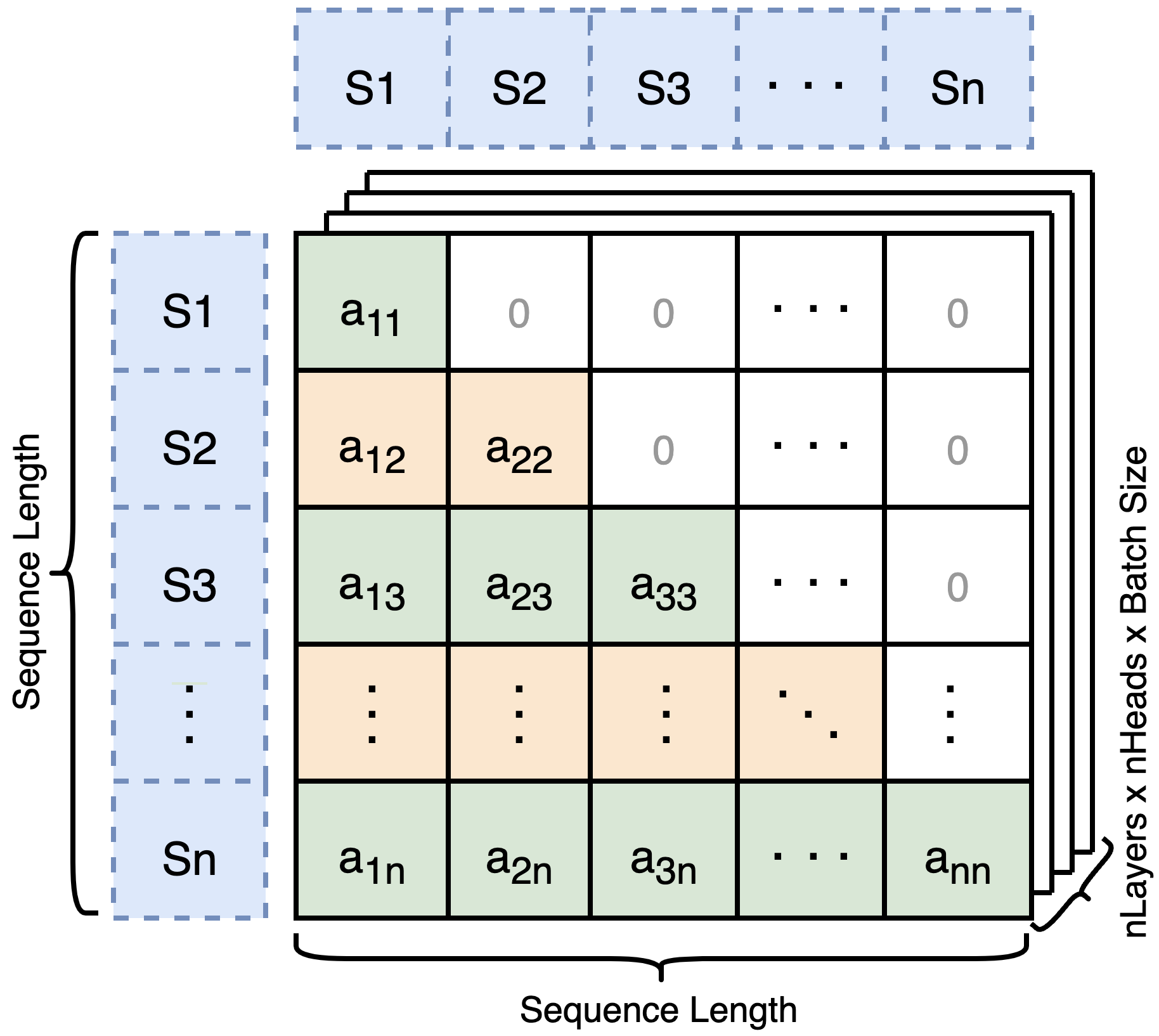}
    \caption{\textmd{Self-Attention States: $a_{ij}$ represents the attention given to the $i$-th token when processing or generating the $j$-th token. Causal masking prevents attention to future tokens by setting their attention weight to zero.}}
    \label{fig:attn}
\end{figure}

In the context of this equation and referring to Figure \ref{fig:attn}, where $a_{ij}$ denotes the attention given to token $i$ when referencing token $j$, each row in the figure sums to 1. This summation indicates that the attention weights assigned to all tokens preceding (and including) token $j$ when referencing $j$ imply a probability distribution over the tokens.  We use KLD loss over this probability distribution as depicted in the loss term $\mathcal{L}_{\text{attn, t}}$~\eqref{eq:attn}, representing the loss from a single row in Figure \ref{fig:attn}. The causal mask in Figure \ref{fig:attn} zeroes out future positions in the input sequence. This allows each token to attend only to previous tokens, ensuring that the model generates output based only on current information during generation or decoding. This maintains causality and prevents leakage from future tokens.

\begin{gather}
\mathcal{L}_{\text{attn, t}} = \sum_{i=1}^{t} a_{it, teacher} \log \left( \frac{a_{it, teacher}}{a_{it, student}} \right)
\label{eq:attn}
\end{gather}

The loss term $\mathcal{L}_{\text{attn}}$ in equation ~\eqref{eq:kd} is obtained by normalizing the sum of $\mathcal{L}_{\text{attn, t}}$ across the entire sequence length (representing all rows in Figure \ref{fig:attn}), batch size, number of attention heads, and layers in the architecture.

\begin{gather}
\mathcal{L}_{\text{KD}} = \mathcal{L}_{\text{pred}} + \mathcal{L}_{\text{attn}}
\label{eq:kd}
\end{gather}

In our ablation study, we separately explore prediction layer distillation~\eqref{eq:pred}, attention based distillation~\eqref{eq:attn}, and combined distillation techniques~\eqref{eq:kd}. Our findings indicate that combined distillation emerges as the most effective method for enhancing the instruction-following capability of the student model. Prediction layer distillation plays a pivotal role in achieving this outcome. Additionally, attention based distillation offers guidance in the distillation process, yielding better results than either standalone distillation method. Although we do not delve into distillation for pre-training in this study, our observations suggest that attention based distillation could potentially prove crucial during that phase of training.

We leverage Mistral 7B v0.1 Base \cite{jiang2023mistral} as our student model and conduct ablations to validate the aforementioned theory. For comparison, Mistral 7B Instruct v2 is employed as a teacher model of similar size, while Mixtral 8x7B Instruct serves as a larger-scale teacher model. These ablations aim to provide empirical evidence supporting the theoretical framework. Subsequently, upon establishing the foundation, we scale up our student model by merging two Mistral models into a 2x7B (13B) mixture of experts \label{sec:{2x7B}} \cite{goddard2024arcee} while using Mixtral 8x7B Instruct as the teacher model. An intriguing observation emerged post training: while our student model may not fully reach the performance level of the teacher model, it exhibits alignment, stemming from the teacher model's adherence to human preferences. Any further effort to enhance alignment using DPO led to a decrease in evaluation metrics. This could indicate that the initial alignment achieved post-training is robust, and further alignment efforts may require an iterative approach with significantly improved preference data and lowered learning rates.

\vspace{-0.15cm}

\subsection{D\MakeLowercase{omain} A\MakeLowercase{lignment from} E\MakeLowercase{xpert}}

Further training an instruction tuned and aligned model in a given domain causes the model to lose its generalising ability; while the model excels at in-domain tasks, it under performs on unseen and out of domain tasks. 

Domain Alignment from Expert (DAE) uses the above theory as foundation and presents as a novel approach to impart domain knowledge to the trained and aligned model while controlling its generalisation capability. Though we experimented DAE extensively on the student model we got in the previous training phase of knowledge distillation (KD), we successfully conducted an experiment to align a Mistral 7B Instruct v2 as well, implying that this approach could yield good results for any model.

In training, the language model $q_{\theta}$ is initialized from the student knowledge distilled model from the previous training phase. We also employ a base reference model $p_{\text{ref}}$ derived from the same knowledge distilled model. The reference model serves to prevent the model from deviating too far from its generalized state during training. Additionally, the domain expert ($p_{\text{e}}$) is a model specifically trained and aligned on domain-specific data. It is crucial for this expert to excel within its designated domain without the necessity to generalize to unseen or out-of-domain tasks. Within a training batch, the student model utilizes the domain expert as the teacher for in-domain samples and the reference model for non-domain samples. Our study demonstrates that even with domain data just being 10\% of the total training data, the model can effectively learn about the domain while still maintaining generalizability.

\begin{figure}[h]
    \centering
    \includegraphics[width = 0.48\textwidth]{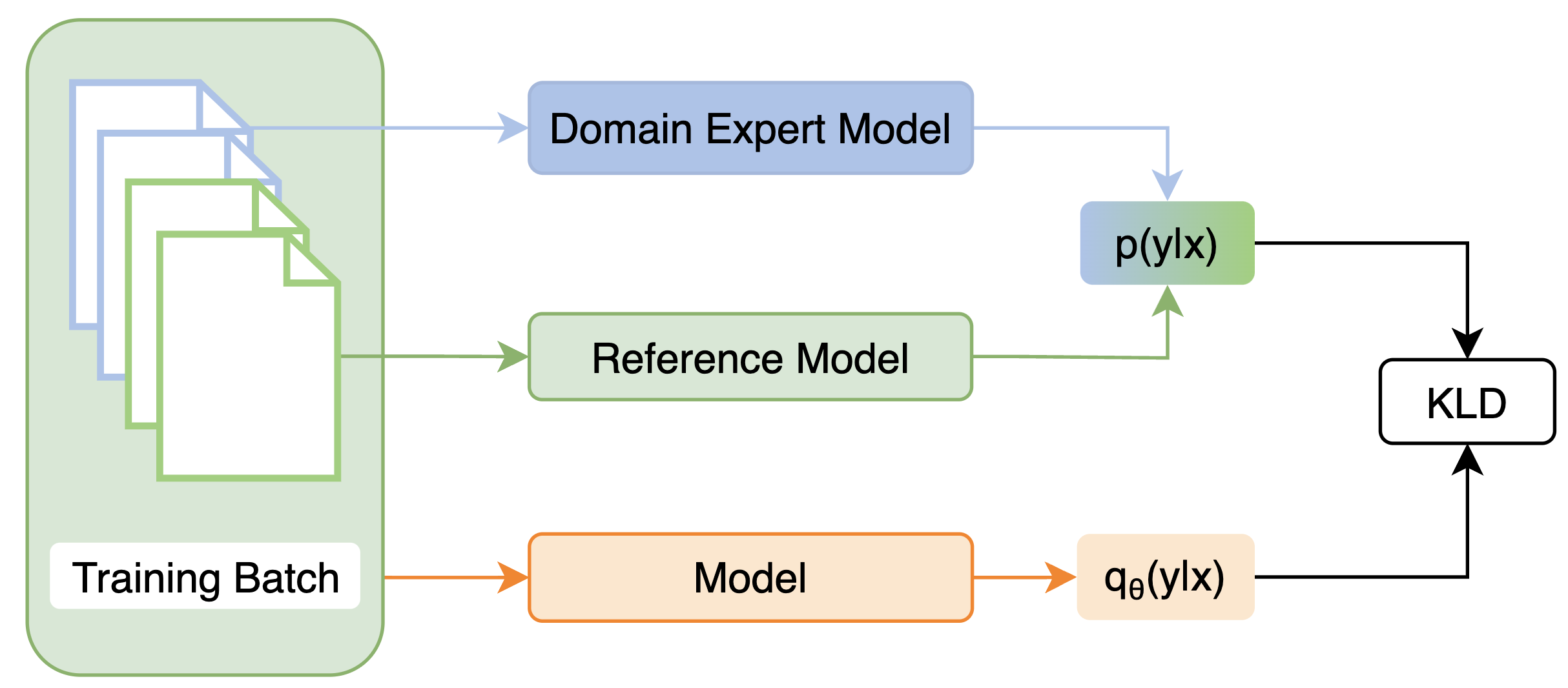}
    \caption{\textmd{DAE: domain samples refer to the domain expert as their teacher, while non-domain samples refer to the reference model. The stacked distributions is considered as the "true" distribution.}}
    \label{fig:dae}
\end{figure}

\vspace{-0.6cm}

\begin{gather}
    \mathcal{L}_{\text{d}} = \frac{1}{T} \sum_{t=1}^{T} \sum_{i=1}^{V} \mathds{1}_{\text{d}}(x) p_{e}(y_i|x_{t' \leq t}) \log \left( \frac{p_{e}(y_i|x_{t' \leq t})}{q_{\theta}(y_i|x_{t' \leq t})} \right) \notag \\[10pt]
    \mathcal{L}_{\text{nd}} = \frac{1}{T} \sum_{t=1}^{T} \sum_{i=1}^{V} \mathds{1}_{\text{nd}}(x) p_{\text{ref}}(y_i|x_{t' \leq t}) \log \left( \frac{p_{\text{ref}}(y_i|x_{t' \leq t})}{q_{\theta}(y_i|x_{t' \leq t})} \right) \notag \\[10pt]
    \mathcal{L}_{\text{DAE}} = \mathcal{L}_{\text{d}} + \mathcal{L}_{\text{nd}} \label{eq:loss_dae}
\end{gather}

where $\mathcal{L}_{\text{d}}$ represents the loss for domain samples that use the domain expert for the true distribution and $\mathcal{L}_{\text{nd}}$ is the loss for general out-of-domain samples that use the reference model. 

With DAE already being computationally intensive due to three models in memory, we opt to use only prediction layer distillation, as attention based distillation entails even higher computational requirements. Additionally, since DAE serves as an alignment phase and operates on an already distilled model, it could potentially function with slightly less guidance during training.

\section{E\MakeLowercase{xperiments}}

In this section, we assess the ability of KD and DAE in training models to perform comparably with the standard SFT and alignment approach. We utilize the Transformers library \cite{wolf-etal-2020-transformers} for supervised fine-tuning and the TRL library \cite{vonwerra2022trl} for DPO alignment atop SFT models.

\begin{figure*}[h]
    \centering
    \includegraphics[width = 1\textwidth]{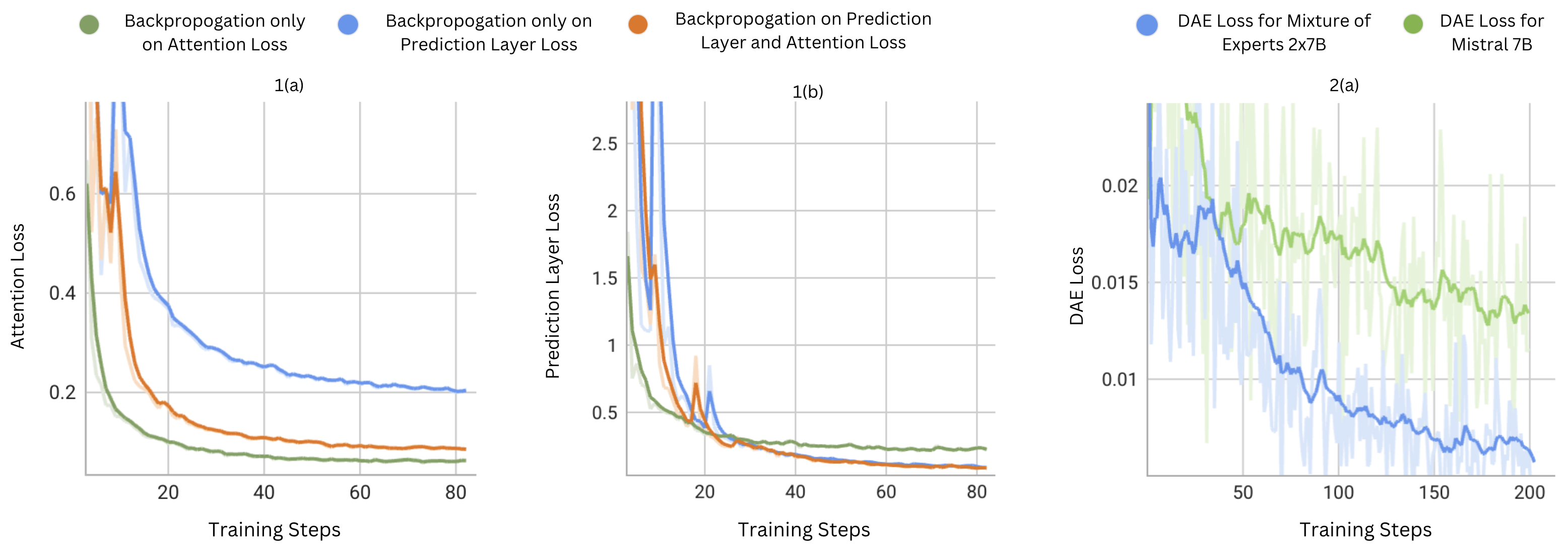}
    \caption{\textmd{Figure 1(a) and 1(b) depict the progression of attention loss and prediction layer loss, respectively. Training is conducted on attention-only loss (green), prediction layer-only loss (blue), and both combined (orange). Despite backpropagation occurring on one or both losses, a natural correlation emerges, along with gradient magnitudes aligning with expected trends. In Figure 2(a), we observe the Domain Alignment from Expert (DAE) training loss on Mistral 7B (green) and 2x7B Mixture of Expert (blue). Notably, the loss curve for MoE 2x7B is significantly lower than that of Mistral 7B, as anticipated.}}
    \label{fig:train_curve}
\end{figure*}

\subsection{Models}

\subsubsection{Ablation Study}
\label{sec:ablation}
We commence with several ablations on Mistral 7B v0.1 Base to substantiate our theory. Mistral 7B v0.1 Base is employed as the student model, while Mistral 7B Instruct v2 and Mixtral 8x7B v0.1 Instruct serve as teachers for Knowledge Distillation (KD) experiments. These experiments were benchmarked,

alongside the SFT experiment, on 20k samples sourced from the public dataset Ultrachat \cite{ding2023enhancing}.

\subsubsection{KD at Scale}
\label{sec:kds}
Following successful ablations, we expand these experiments to include 7B and 2x7B models as students with Mixtral 8x7B v0.1 Instruct as the teacher. We merge two Mistral 7B models using mergekit \cite{goddard2024arcee} to get a 2x7B MoE base model.

\subsubsection{Domain Alignment from Expert}
\label{sec:dae}
After successful scaling, we proceed to align the scaled 2x7B model within the e-commerce domain. Additionally, we apply the same approach to Mistral 7B Instruct v0.1 to verify the effectiveness of DAE on any model. To establish an e-commerce domain expert, we train a Mistral 7B v0.2 Base model on internal e-commerce data using SFT and DPO. While this model performs well on our internal e-commerce evaluation, its performance on the public evaluation is average.

\subsection{Dataset}

We utilize a different mix of public and e-commerce datasets at each training stage. Our emphasis is not primarily on the quality of the datasets but rather on ensuring that the mix is a fair and diverse representation of all domains.

\begin{itemize}
    \item For \ref{sec:ablation}, we utilize 20k different samples sourced from the public dataset Ultrachat \cite{ding2023enhancing}.
    
    \item For \ref{sec:kds}, we utilize numerous single-turn and multi-turn public datasets \cite{cobbe2021gsm8k, yu2023metamath, ding2023enhancing, SlimOrca, codealpaca, zhou2023lima} across diverse domains such as mathematics, coding, reasoning, and Chain-of-Thought.
    
    \item For \ref{sec:dae}, we use 45k samples from the \ref{sec:kds} dataset and add 5k samples from our in-house e-commerce dataset.
    
    \item Our in-house e-commerce dataset is a GPT-4 generated instruction dataset spanning across various e-commerce tasks such as extraction, reasoning, QnA and summarisation.
\end{itemize}

\subsection{Training Configuration}

Deriving from the training configurations of supervised fine-tuning and alignment from Llama 2 \cite{touvron2023llama} and Zephyr \cite{tunstall2023zephyr}, we adopt similar configurations. For \ref{sec:ablation} and \ref{sec:kds}, we adhere to configurations similar to SFT and we adopt the DPO configurations for \ref{sec:dae}.

\begin{itemize}
    \item For \ref{sec:ablation}, we use a cosine learning rate scheduler with a peak learning rate of 2e-5 and 10\% warmup steps. We train for 2 epochs with a global batch size of 512 and packing with a sequence length of 2048 tokens.
    
    \item For \ref{sec:kds}, we use a cosine learning rate scheduler with a peak learning rate of 2e-5 and 10\% warmup steps. We train for 1 epoch with a global batch size of 6000 and packing with a sequence length of 2048 tokens.
    
    \item For \ref{sec:dae}, we use a constant learning rate scheduler with a peak learning rate of 5e-7 and 10\% warmup steps. We train for 1 epoch with a global batch size of 64 and packing with a sequence length of 4096 tokens.
\end{itemize}

\subsection{Leaderboard Evaluation}

Our main evaluations are on single-turn and multi-turn chat benchmarks that measure a model’s ability to follow instructions and respond to challenging prompts across a diverse range of domains. We mainly use the following two benchmarks in our study:

    \begin{table*}[h!]
    \centering
    \begin{threeparttable}
    \begin{tabular}{l|cccc}
    \hline
    \textbf{Model}          & \textbf{Size} & \textbf{MT-Bench (score)} & \textbf{AlpacaEval (win \%)} & \textbf{EcommEval (score)} \\ \hline
    MPT-Chat                & 7B            & 5.42                      & -                     & -                           \\
    Zephyr            & 7B            & 7.34                     & 90.60                        & -                   \\
    Mistral-Instruct v0.2   & 7B            & 7.6                       & -                     & 8.45                           \\  \hline
    \textbf{Flip-7B-Instruct--\bm{$\alpha$}}         & 7B            & \textbf{7.58}             & \textbf{90.17}                     & 8.36              \\ \hline
    \textbf{Flip-7B-Instruct-\bm{$\beta$}}         & 7B            & \textbf{7.70}             & \textbf{91.60}                     & 8.75             \\ \hline
    Falcon-Instruct         & 40B           & 5.17                      & 45.71                      & -                       \\
    Guanaco                 & 65B           & 6.41                      & 71.80                      & -                       \\
    Llama2-Chat             & 70B           & 6.86                      & 92.66                      & 7.93                       \\
    Vicuna v1.3             & 33B           & 7.12                      & 88.99                      & -                       \\
    WizardLM v1.0           & 70B           & 7.71                      & -                       & -                          \\ \hline               
    \textbf{Flip-2x7B-Instruct-\bm{$\alpha$}}         & 13B            & \textbf{7.70}             & \textbf{92.67}                      & 8.4              \\ \hline
    \textbf{Flip-2x7B-Instruct-\bm{$\beta$}}         & 13B            & \textbf{7.90}             & \textbf{93.04}                      & 8.95              \\ \hline
    Mixtral 8x7B Instruct v0.1               & 45B             & 8.30                      & 94.80                      & 9.05                       \\
    GPT-3.5-turbo           & -             & 7.94                      & 89.37                      & 9.17                       \\
    Claude 1                & -             & 7.90                      & 88.39                      & -                       \\
    Claude 2                & -             & 8.06                      & 91.36                      & -                       \\
    GPT-4                   & -             & \textbf{8.99}                      & \textbf{95.28}                      & 9.64              \\ \hline
    \end{tabular}
    \caption{\textmd{Evaluation results for open-access and proprietary models on MT-Bench, AlpacaEval and our in-house EcommEval. A dash (-) indicates information or an evaluation that is not publicly available.}}
    \label{tab:model_comparison}
    \end{threeparttable}
\end{table*}

\begin{table}[h!]
    \centering
    \begin{threeparttable}
    \begin{tabular}{l|cccc}
    \hline
    \textbf{Experiment} & \textbf{mean} & \textbf{std} \\ \hline
    Random Samples (R) + SFT                & 5.77            & 0.368                          \\
    QA Samples + SFT            & 6.22            & 0.123                  \\
    R + KD(w/o pred): Mistral Instruct v2 & 6.53 & 0.032 \\
    R + KD(w/o attn): Mistral Instruct v2 & 6.71 & 0.047 \\
    R + KD: Mistral Instruct v2   & 7.05            & 0.051   \\
    R + KD: Mixtral 8x7B Instruct         & 7.24            & 0.105 \\ \hline
    \end{tabular}
    \caption{\textmd{Evaluation results of ablation study (\ref{sec:ablation}) on 20k random (R) and quality assured (QA) samples from ultrachat: mean and standard deviation of MT-Bench scores are calculated from five runs for each experiment (row).}}
    \label{tab:ablation_model_comparison}
    \end{threeparttable}
\vspace{-0.55cm}
\end{table}

\begin{itemize}
    \item \textbf{MT-Bench} \cite{zheng2023judging}: A multi-turn evaluation benchmark comprising 80 multi-turn questions across 8 different domains, namely, Writing, Roleplay, Reasoning, Math, Code, STEM, Humanities, and Extraction. For each of the 80 questions, the model must provide single-turn responses and then respond to a predefined follow-up question. The model results are scored using GPT-4 on a scale of 1 to 10, with the final score being an average over the two turns.

    \item \textbf{AlpacaEval} \cite{alpaca_eval}: A single-turn benchmark where a model is tasked with generating responses to 805 questions covering various topics, predominantly focused on helpfulness. Models are scored by GPT-4, and the final metric is the pairwise win-rate against a baseline model, text-davinci-003.

    \item \textbf{EcommEval}: An elaborate single-turn in-house benchmark comprising 2000 questions across 20 e-commerce tasks. The model is scored by GPT4-turbo\footnote{https://platform.openai.com/docs/models/gpt-4-and-gpt-4-turbo} on a scale of 1 to 10 under multiple aspects such as instruction following, factuality, accuracy, and helpfulness.
\end{itemize}

\section{R\MakeLowercase{esults}}

We present the results of our ablation study (\ref{sec:ablation}) in Table \ref{tab:ablation_model_comparison}. We conducted experiments on 20,000 random (R) and quality assured (QA) samples sourced from the ultrachat \cite{ding2023enhancing} dataset. Specifically, our experiments encompassed SFT on random samples (row 1), SFT on QA samples (row 2), and Knowledge Distillation employing Mistral Instruct v2 and Mixtral 8x7B Instruct as teachers, with Mistral 7B v0.1 Base acting as the student (rows 3 to 6). For KD, we run separate experiments for attention based distillation [see Eq. ~\eqref{eq:attn}], prediction layer distillation [see Eq.~\eqref{eq:pred}], and combined distillation [see Eq.~\eqref{eq:kd}], as illustrated in Figure \ref{fig:train_curve}. We used MT Bench scores to benchmark these experiments. Our analysis indicates that SFT yields results characterized by a lower mean score and higher deviation, while experiments employing KD exhibit promising outcomes, generally evidenced by higher mean scores and lower deviations, indicating potential for improved performance. We used MT Bench scores to benchmark these experiments.

% \vspace{-0.55cm}

We introduce four models derived from our final scaled-up experiments in table \ref{tab:model_comparison}.

\begin{itemize}
    \item \textbf{Flip-7B-Instruct-\bm{$\alpha$}}: We run KDS (\ref{sec:kds}) with Mistral 7B v0.1 Base as the student and Mixtral 8x7B Instruct as the teacher. We observe an MT-Bench score of 7.58, 90.17\% win-rate in AlpacaEval, and 8.36 on EcommEval.
    
    \item \textbf{Flip-7B-Instruct-\bm{$\beta$}}: We run e-commerce DAE (\ref{sec:dae}) on Flip-7B-Instruct-$\alpha$ as the student, with our e-commerce model serving as the domain expert. This model achieves an MT-Bench score of 7.69, 91.6\% win-rate in AlpacaEval, and 8.75 on EcommEval.

    \item \textbf{Flip-2x7B-Instruct-\bm{$\alpha$}}: We run KDS (\ref{sec:kds}) with Mistral 2x7B as the student and Mixtral 8x7B Instruct as the teacher. We observe an MT-Bench score of 7.7, 92.67\% win-rate in AlpacaEval, and 8.4 on EcommEval.
    
    \item \textbf{Flip-2x7B-Instruct-\bm{$\beta$}}: We run e-commerce DAE (\ref{sec:dae}) on Flip-2x7B-Instruct-$\alpha$ as the student, with our e-commerce model serving as the domain expert. This model achieves an MT-Bench score of 7.90, 93.04\% win-rate in AlpacaEval, and 8.95 on EcommEval.
\end{itemize}

% We have released the alpha versions of the 7B and 2x7B models. However, we are unable to release the beta models due to the proprietary data utilized in training the e-commerce domain expert and its role in the training process of the beta models.

\section{R\MakeLowercase{elated} W\MakeLowercase{ork}}
\textbf{Large Language Models} (LLMs; \cite{thoppilan2022lamda, touvron2023llama}) have opened new avenues in Natural Language Processing (NLP) by solving various complex tasks in a generative manner. Recent works explore instruction tuning \cite{wei2022finetuned, zhang2024instruction} and human preference alignment \cite{li2023reinforcement, rafailov2023direct, hong2024orpo} to create general-purpose safe AI assistants. Instruction tuning enables the model to follow instructions and excel at unseen tasks, while alignment step aligns the model with ethical and societal principles. Moreover, the open-sourcing of LLM architectures such as Mistral \cite{jiang2023mistral} and Llama \cite{touvron2023llama} plays a pivotal role in research efforts, and also serves as the foundation of our work.

\textbf{Knowledge Distillation} (KD; \cite{hinton2015distilling}) aims to teach a smaller student model by transferring knowledge from a larger teacher model. This approach popularly has been looked at as a method to compress large models for the ease of deployment. KD has shown great promise in text classification tasks \cite{song2020lightpaff, liang2021mixkd} by learning from the output distribution of the teacher model, or by learning from the teacher's self-attention output \cite{wang2021minilmv2}. In text generation tasks the student learns by matching the output distribution of the teacher model at every time-step \cite{gu2023minillm, sanh2020distilbert} or by a supervised fine-tuning on teacher outputs \cite{xu2023wizardlm, wang2023selfinstruct, alpaca}. 

\section{C\MakeLowercase{onclusion}}

In this work, we investigate the problem of knowledge distillation at scale, leveraging teachers of varying sizes. Our KDS approach employs a larger LLM, Mixtral 8x7B Instruct, to teach smaller models of 7B and 2x7B (13B) scale. Conversely, in DAE, we utilize a smaller (7B) domain expert to align larger 2x7B (13B) mixture of experts. We propose knowledge distillation as a legitimate means of model training rather than a mere way to compress larger LLMs for deployment purposes.  

We demonstrate that our models are either comparable to or, in some cases, even outperforms open-access and proprietary much larger LLMs trained using the popular SFT and alignment approach. While we acknowledge the importance of SFT and alignment in training LLMs, we advocate for our KDS approach as a viable training method for models that could benefit from superior teachers. Additionally, our DAE approach proves to be highly effective for domain alignment without compromising generalization. Further research in this direction has the potential to significantly enrich the LLM training toolbox. 

\vspace{0.35cm}
\bibliographystyle{plain}
\renewcommand{\refname}{R\MakeLowercase{eferences}}
\bibliography{main}

\end{document}